\documentclass[10pt,twocolumn,letterpaper]{article}

\usepackage[pagenumbers]{cvpr} % To force page numbers, e.g. for an arXiv version

\usepackage{bbm}
\usepackage{lipsum}
\usepackage{colortbl}
\usepackage{amsthm}

\usepackage[dvipsnames]{xcolor}

\usepackage{multirow}
\usepackage{wasysym}

\usepackage{subcaption}
\usepackage{graphicx}
\usepackage{amsmath}
\usepackage{amssymb}
\usepackage{booktabs}
\usepackage{graphicx}
\usepackage{booktabs} 
\usepackage{enumitem}
\usepackage{multirow}
\usepackage{multicol}
\usepackage{hyperref}

\definecolor{cvprblue}{rgb}{0.21,0.49,0.74}

\def\paperID{1180} % *** Enter the Paper ID here
\def\confName{CVPR}
\def\confYear{2025}

\title{TinyFusion: Diffusion Transformers Learned Shallow}

\author{
    Gongfan Fang\thanks{Equal contribution}, ~Kunjun Li\footnotemark[1], ~Xinyin Ma, ~Xinchao Wang\thanks{Corresponding author} \\
    National University of Singapore \\
    {\small \tt{\{gongfan, kunjun, maxinyin\}@u.nus.edu, xinchao@nus.edu.sg}} \\
}

\begin{document}
\maketitle

\newcommand{\methodname}{TinyDiT}
\newcommand{\graycell}{\cellcolor{gray!15}}
\newcommand{\grayrow}{\rowcolor{gray!15}}
\newcommand{\argmin}{\operatornamewithlimits{argmin}}
\newcommand{\argmax}{\operatornamewithlimits{argmax}}

\begin{abstract}
Diffusion Transformers have demonstrated remarkable capabilities in image generation but often come with excessive parameterization, resulting in considerable inference overhead in real-world applications. In this work, we present TinyFusion, a depth pruning method designed to remove redundant layers from diffusion transformers via end-to-end learning. The core principle of our approach is to create a pruned model with high \emph{recoverability}, allowing it to regain strong performance after fine-tuning. To accomplish this, we introduce a differentiable sampling technique to make pruning learnable, paired with a co-optimized parameter to simulate future fine-tuning. While prior works focus on minimizing loss or error after pruning, our method explicitly models and optimizes the post-fine-tuning performance of pruned models. Experimental results indicate that this learnable paradigm offers substantial benefits for layer pruning of diffusion transformers, surpassing existing importance-based and error-based methods. Additionally, TinyFusion exhibits strong generalization across diverse architectures, such as DiTs, MARs, and SiTs. Experiments with DiT-XL show that TinyFusion can craft a shallow diffusion transformer at less than 7\% of the pre-training cost, achieving a 2$\times$ speedup with an FID score of 2.86, outperforming competitors with comparable efficiency. Code is available at \url{https://github.com/VainF/TinyFusion}
\end{abstract}

\section{Introduction}

Diffusion Transformers have emerged as a cornerstone architecture for generative tasks, achieving notable success in areas such as image~\cite{peebles2023scalable,esser2024scaling,flux} and video synthesis~\cite{opensora,pku_yuan_lab_and_tuzhan_ai_etc_2024_10948109}. This success has also led to the widespread availability of high-quality pre-trained models on the Internet, greatly accelerating and supporting the development of various downstream applications~\cite{yang2023diffusion,chen2023pixartalpha,gao2023editanything,yu2023inpaint}. 
However, pre-trained diffusion transformers usually come with considerable inference costs due to the huge parameter scale, which poses significant challenges for deployment. To resolve this problem, there has been growing interest from both the research community and industry in developing lightweight models~\cite{lin2024sdxl,kim2023bk,fang2023structural,zhao2023mobilediffusion}.

\begin{figure}[t]
    \centering
    \includegraphics[width=\linewidth]{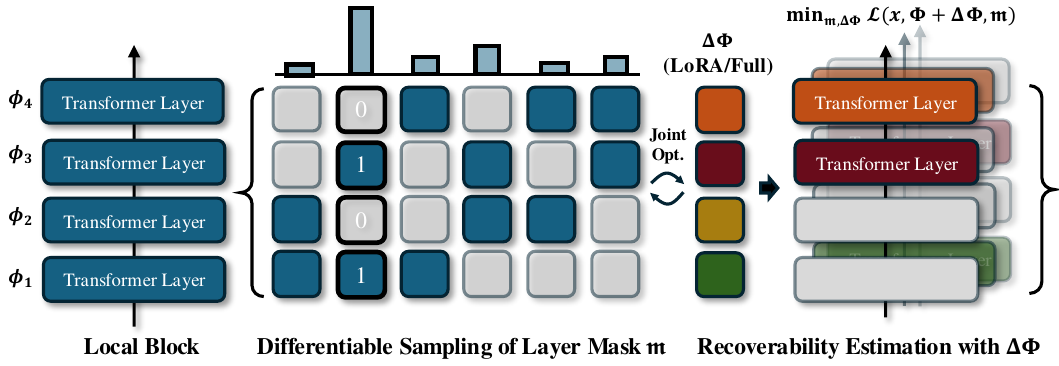}
    \caption{This work presents a learnable approach for pruning the depth of pre-trained diffusion transformers. Our method simultaneously optimizes a differentiable sampling process of layer masks and a weight update to identify a highly recoverable solution, ensuring that the pruned model maintains competitive performance after fine-tuning.}  \label{fig:intro}
\end{figure}

The efficiency of diffusion models is typically influenced by various factors, including the number of sampling steps~\cite{song2020denoising,song2023consistency,lu2022dpm,salimans2022progressive}, operator design~\cite{teng2024dim,dao2022flashattention,xie2024sana}, computational precision~\cite{li2023q,shang2023post,he2024ptqd}, network width~\cite{fang2023structural,castells2024ld} and  depth~\cite{kim2023bk,flux1-lite,men2024shortgpt}.   In this work, we focus on model compression through depth pruning~\cite{yu2022width,men2024shortgpt}, which removes entire layers from the network to reduce the latency. Depth pruning offers a significant advantage in practice: it can achieve a linear acceleration ratio relative to the compression rate on both parallel and non-parallel devices. For example, as will be demonstrated in this work, while 50\% width pruning~\cite{fang2023structural} only yields a 1.6× speedup, pruning 50\% of the layers results in a 2× speedup. This makes depth pruning a flexible and practical method for model compression.

This work follows a standard depth pruning framework: unimportant layers are first removed, and the pruned model is then fine-tuned for performance recovery.  In the literature, depth pruning techniques designed for diffusion transformers or general transformers primarily focus on heuristic approaches, such as carefully designed importance scores~\cite{men2024shortgpt,flux1-lite} or manually configured pruning schemes~\cite{kim2023bk,yu2022width}. These methods adhere to a loss minimization principle~\cite{han2015learning,molchanov2016pruning}, aiming to identify solutions that maintain low loss or error after pruning. This paper investigates the effectiveness of this widely used principle in the context of depth compression. Through experiments, we examined the relationship between calibration loss observed post-pruning and the performance after fine-tuning. This is achieved by extensively sampling 100,000 models via random pruning, exhibiting different levels of calibration loss in the searching space. Based on this, we analyzed the effectiveness of existing pruning algorithms, such as the feature similarity~\cite{flux1-lite,men2024shortgpt} and sensitivity analysis~\cite{han2015learning}, which indeed achieve low calibration losses in the solution space. However, the performance of all these models after fine-tuning often falls short of expectations. This indicates that the loss minimization principle may not be well-suited for diffusion transformers.

Building on these insights, we reassessed the underlying principles for effective layer pruning in diffusion transformers. Fine-tuning diffusion transformers is an extremely time-consuming process. Instead of searching for a model that minimizes loss immediately after pruning, we propose identifying candidate models with strong recoverability, enabling superior post-fine-tuning performance. Achieving this goal is particularly challenging, as it requires the integration of two distinct processes, pruning and fine-tuning, which involve non-differentiable operations and cannot be directly optimized via gradient descent. 

To this end, we propose a learnable depth pruning method that effectively integrates pruning and fine-tuning. As shown in Figure~\ref{fig:intro}, we model the pruning and fine-tuning of a diffusion transformer as a differentiable sampling process of layer masks~\cite{gumbel1954statistical,jang2016categorical,fang2024maskllm}, combined with a co-optimized weight update to simulate future fine-tuning. Our objective is to iteratively refine this distribution so that networks with higher recoverability are more likely to be sampled. This is achieved through a straightforward strategy: if a sampled pruning decision results in strong recoverability, similar pruning patterns will have an increased probability of being sampled. This approach promotes the exploration of potentially valuable solutions while disregarding less effective ones. Additionally, the proposed method is highly efficient, and we demonstrate that a suitable solution can emerge within a few training steps.

To evaluate the effectiveness of the proposed method, we conduct extensive experiments on various transformer-based diffusion models, including DiTs~\cite{peebles2023scalable}, MARs~\cite{li2024autoregressive}, SiTs~\cite{ma2024sit}. The learnable approach is highly efficient. It is able to identify redundant layers in diffusion transformers with 1-epoch training on the dataset, which effectively crafts shallow diffusion transformers from pre-trained models with high recoverability. For instance, while the models pruned by TinyFusion initially exhibit relatively high calibration loss after removing 50\% of layers, they recover quickly through fine-tuning, achieving a significantly more competitive FID score (5.73 vs. 22.28) compared to baseline methods that only minimize immediate loss, using just 1\% of the pre-training cost. Additionally, we also explore the role of knowledge distillation in enhancing recoverability~\cite{hinton2015distilling,kim2023bk} by introducing a MaskedKD variant. MaskedKD mitigates the negative impact of the massive or outlier activations~\cite{sun2024massive} in hidden states, which can significantly affect the performance and reliability of fine-tuning. With MaskedKD, the FID score improves from 5.73 to 3.73 with only 1\% of pre-training cost. Extending the training to 7\% of the pre-training cost further reduces the FID to 2.86, just 0.4 higher than the original model with doubled depth.

Therefore, the main contribution of this work lies in a learnable method to craft shallow diffusion transformers from pre-trained ones, which explicitly optimizes the recoverability of pruned models. The method is general for various architectures, including DiTs, MARs and SiTs.

\begin{figure*}[t]
    \centering
    \includegraphics[width=0.85\linewidth]{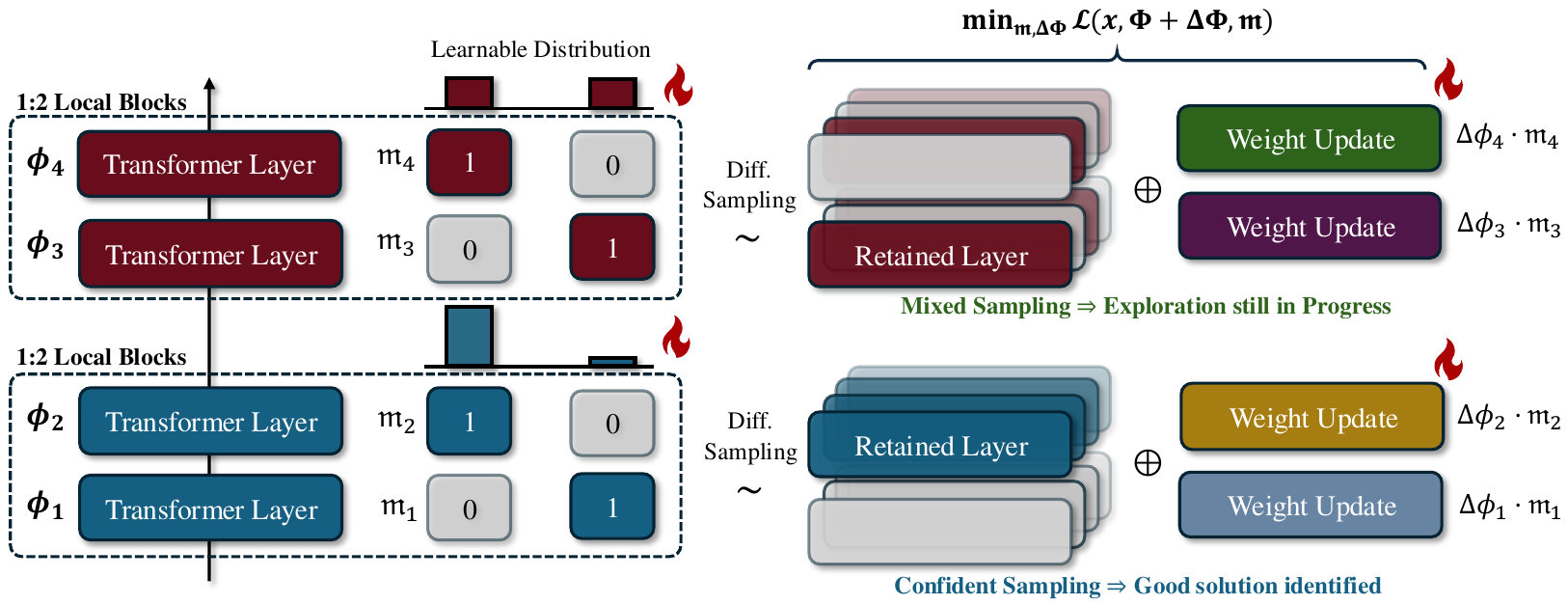}
    \caption{The proposed TinyFusion method learns to perform a differentiable sampling of candidate solutions, jointly optimized with a weight update to estimate recoverability. This approach aims to increase the likelihood of favorable solutions that ensure strong post-fine-tuning performance. After training, local structures with the highest sampling probabilities are retained.}
    \label{fig:framework}
\end{figure*}

\section{Related Works}

\paragraph{Network Pruning and Depth Reduction.} Network pruning is a widely used approach for compressing pre-trained diffusion models by eliminating redundant parameters~\cite{fang2023structural,castells2024ld,wang2024sparsedm, li2024snapfusion}. Diff-Pruning~\cite{fang2023structural} introduces a gradient-based technique to streamline the width of UNet, followed by a simple fine-tuning to recover the performance. SparseDM~\cite{wang2024sparsedm} applies sparsity to pre-trained diffusion models via the Straight-Through Estimator (STE)~\cite{bengio2013estimating}, achieving a 50\% reduction in MACs with only a 1.22 increase in FID on average. While width pruning and sparsity help reduce memory overhead, they often offer limited speed improvements, especially on parallel devices like GPUs. Consequently, depth reduction has gained significant attention in the past few years, as removing entire layers enables better speedup proportional to the pruning ratio~\cite{yu2022width,men2024shortgpt,kim2024shortened,leedit,zhao2023mobilediffusion, zhang2024laptop, lee2023koala}. Adaptive depth reduction techniques, such as MoD~\cite{raposo2024mixture} and depth-aware transformers~\cite{elbayad2019depth}, have also been proposed. Despite these advances, most existing methods are still based on empirical or heuristic strategies, such as carefully designed importance criteria~\cite{men2024shortgpt,yu2022width}, sensitivity analyses~\cite{han2015learning} or manually designed schemes~\cite{kim2023bk}, which often do not yield strong performance guarantee after fine-tuning.

%It is still unclear how effective these methods 
%reduction remains challenging, particularly in identifying non-essential layers that consider fine-tuning. Current methods primarily rely on empirical criteria~\cite{men2024shortgpt,yu2022width, kim2023bk} and sensitivity analyses~\cite{han2015learning}. There is no strong guarantee about the performance after fine-tuning.

\paragraph{Efficient Diffusion Transformers.} 
Developing efficient diffusion transformers has become an appealing focus within the community, where significant efforts have been made to enhance efficiency from various perspectives, including linear attention mechanisms~\cite{fei2024dimba,teng2024dim,xie2024sana}, compact architectures~\cite{tian2024u}, non-autoregressive transformers~\cite{chang2022maskgit,tian2024visual,ni2024revisiting,fei2024scaling}, pruning~\cite{kim2023bk,fang2023structural}, quantization~\cite{li2023q,shang2023post,he2024ptqd}, feature caching~\cite{ma2024learningtocache,zhao2024real}, etc. In this work, we focus on compressing the depth of pre-trained diffusion transformers and introduce a learnable method that directly optimizes recoverability, which is able to achieve satisfactory results with low re-training costs.

\newcommand{\layer}{\boldsymbol{\phi}}
\newcommand{\mask}{\boldsymbol{\frak{m}}}
\section{Method}

\subsection{Shallow Generative Transformers by Pruning}

This work aims to derive a shallow diffusion transformer by pruning a pre-trained model. For simplicity, all vectors in this paper are column vectors. Consider a $L$-layer transformer, parameterized by $\Phi_{L\times D}=\left[\layer_1, \layer_2, \cdots, \layer_L\right]^{\intercal}$, where each element $\layer_i$ encompasses all learnable parameters of a transformer layer as a $D$-dim column vector, which includes the weights of both attention layers and MLPs. Depth pruning seeks to find a binary layer mask $\mask_{L\times 1}=\left[m_1, m_2, \cdots, m_L\right]^\intercal$, that removes a layer by:
\begin{equation}
    x_{i+1} = m_i \layer_i(x_i) + (1-m_i)x_i = 
    \begin{cases}
        \layer_i(x_i), \; & \text{if}\ m_i=1, \\
        x_i, \; & \text{otherwise}, \\
    \end{cases}
    \label{eqn:forward}
\end{equation}
where the $x_i$ and $\layer_i(x_i)$ refers to the input and output of layer $\layer_i$. To obtain the mask, a common paradigm in prior work is to minimize the loss $\mathcal{L}$ after pruning, which can be formulated as $\min_{\mask} \mathbb{E}_x \left[ \mathcal{L}(x, \Phi, \mask) \right]$. However, as we will show in the experiments, this objective -- though widely adopted in discriminative tasks -- may not be well-suited to pruning diffusion transformers. Instead, we are more interested in the recoverability of pruned models. To achieve this, we incorporate an additional weight update into the optimization problem and extend the objective by:
\begin{equation}
    \min_{\mask}\underbrace{\min_{\Delta\Phi} \mathbb{E}_x \left[ \mathcal{L}(x, \Phi+\Delta\Phi, \mask)\right]}_{\textit{Recoverability: Post-Fine-Tuning Performance}}, \label{eqn:objective}
\end{equation}
where $\Delta\Phi=\{\Delta\layer_1, \Delta\layer_2, \cdots, \Delta\layer_M\}$ represents appropriate update from fine-tuning. The objective formulated by Equation~\ref{eqn:objective} poses two challenges: 1) The non-differentiable nature of layer selection prevents direct optimization using gradient descent; 2) The inner optimization over the retained layers makes it computationally intractable to explore the entire search space, as this process necessitates selecting a candidate model and fine-tuning it for evaluation. To address this, we propose TinyFusion that makes both the pruning and recoverability optimizable.

\subsection{TinyFusion: Learnable Depth Pruning}

\paragraph{A Probabilistic Perspective.}  This work models Equation~\ref{eqn:objective} from a probabilistic standpoint. We hypothesize that the mask $\mask$ produced by ``ideal'' pruning methods (might be not unique) should follow a certain distribution. To model this, it is intuitive to associate every possible mask $\mask$ with a probability value $p(\mask)$, thus forming a categorical distribution. Without any prior knowledge, the assessment of pruning masks begins with a uniform distribution. However, directly sampling from this initial distribution is highly inefficient due to the vast search space. For instance, pruning a 28-layer model by 50\% involves evaluating $\binom{28}{14} = 40,116,600$ possible solutions. To overcome this challenge, this work introduces an advanced and learnable algorithm capable of using evaluation results as feedback to iteratively refine the mask distribution. The basic idea is that if certain masks exhibit positive results, then other masks with \emph{similar pattern} may also be potential solutions and thus should have a higher likelihood of sampling in subsequent evaluations, allowing for a more focused search on promising solutions. However, the definition of ``similarity pattern'' is still unclear so far.

\paragraph{Sampling Local Structures.} \label{sec:local_structure} In this work, we demonstrate that local structures, as illustrated in Figure~\ref{fig:framework}, can serve as effective anchors for modeling the relationships between different masks. If a pruning mask leads to certain local structures and yields competitive results after fine-tuning, then other masks yielding the same local patterns are also likely to be positive solutions. This can be achieved by dividing the original model into $K$ non-overlapping blocks, represented as $ \Phi = \left[ \Phi_1, \Phi_2, \cdots, \Phi_K \right]^{\intercal}$. For simplicity, we assume each block $\Phi_k=\left[\phi_{k1}, \phi_{k2}, \cdots, \phi_{kM}\right]^{\intercal}$ contains exactly $M$ layers, although they can have varied lengths. Instead of performing global layer pruning, we propose an N:M scheme for local layer pruning, where, for each block $\Phi_k$ with $M$ layers, $N$ layers are retained. This results in a set of local binary masks $\mask=[\mask_1, \mask_2, \ldots, \mask_K]^{\intercal}$. Similarly, the distribution of a local mask $\mask_k$ is modeled using a categorical distribution $p(\mask_k)$. We perform independent sampling of local binary masks and combine them for pruning, which presents the joint distribution:
\begin{equation}
    p(\mask) = p(\mask_1) \cdot p(\mask_2) \cdots p(\mask_K)
\end{equation}
If some local distributions $p(\mask_k)$ exhibit high confidence in the corresponding blocks, the system will tend to sample those positive patterns frequently and keep active explorations in other local blocks. Based on this concept, we introduce differential sampling to make the above process learnable.

\paragraph{Differentiable Sampling.} Considering the sampling process of a local mask $\mask_k$, which corresponds a local block $\Phi_k$ and is modeled by a categorical distribution $p(\mask_k)$. With the N:M scheme, there are $\binom{M}{N}$ possible masks. We construct a special matrix $\hat{\mask}^{N:M}$ to enumerate all possible masks. For example, 2:3 layer pruning will lead to the candidate matrix $\hat{\mask}^{2:3}=\left[ \left[1,1,0\right],\left[1,0,1\right],\left[0,1,1\right]\right]$. In this case, each block will have three probabilities $p(\mask_k)=\left[p_{k1}, p_{k2}, p_{k3}\right]$. For simplicity, we omit $\mask_k$ and $k$ and use $p_i$ to represent the probability of sampling $i$-th element in $\hat{\mask}^{N:M}$. A popular method to make a sampling process differentiable is Gumbel-Softmax~\cite{jang2016categorical,gumbel1954statistical,fang2024maskllm}:
\begin{equation}
    y = \text{one-hot} \left(\frac{\exp( (g_i + \log p_i)/\tau)}{\sum_j \exp( (g_j + \log p_j)/\tau)}\right). \label{eqn:gumbel_softmax}
\end{equation}
where $g_i$ is random noise drawn from the Gumbel distribution $\textit{Gumbel}(0, 1)$ and $\tau$ refers to the temperature term. The output $y$ is the index of the sampled mask. Here a Straight-Through Estimator \cite{bengio2013estimating} is applied to the one-hot operation, where the onehot operation is enabled during forward and is treated as an identity function during backward. Leveraging the one-hot index $y$ and the candidate set $\hat{\mask}^{N:M}$, we can draw a mask $\mask\sim p(\mask)$ through a simple index operation:
\begin{equation}
    \mask = y^{\intercal} \hat{\mask}
\end{equation}
Notably, when $\tau \rightarrow 0$, the STE gradients will approximate the true gradients, yet with a higher variance which is negative for training~\cite{jang2016categorical}. Thus, a scheduler is typically employed to initiate training with a high temperature, gradually reducing it over time.

\begin{figure}[t]
    \centering
    \includegraphics[width=0.89\linewidth]{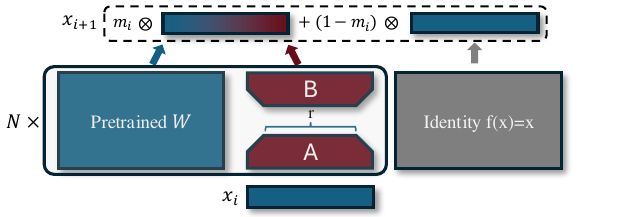}
    \caption{An example of forward propagation with differentiable pruning mask $m_i$ and LoRA for recoverability estimation. }
    \label{fig:lora}
\end{figure}

\begin{table*}[t]
    \centering
    \small
    %\resizebox{\linewidth}{!}{
    \begin{tabular}{l r r r | c c c c c | c}
    \toprule
     \bf Method  & \bf Depth & \bf \#Param & \bf Iters & \bf IS $\uparrow$ & \bf FID $\downarrow$ & \bf sFID $\downarrow$ & \bf Prec. $\uparrow$ & \bf Recall $\uparrow$ & \bf Sampling it/s $\uparrow$ \\
    \midrule
     DiT-XL/2~\cite{peebles2023scalable} & 28 & 675 M & 7,000 K & 278.24 & 2.27 & 4.60 & 0.83 & 0.57 & 6.91 \\
     DiT-XL/2~\cite{peebles2023scalable} & 28 & 675 M & 2,000 K & 240.22 & 2.73 & 4.46 & 0.83 & 	0.55 & 6.91 \\
     DiT-XL/2~\cite{peebles2023scalable} & 28 & 675 M & 1,000 K & 157.83 &	5.53 &	4.60	& 0.80 & 0.53 & 6.91 \\
     U-ViT-H/2~\cite{bao2023all} & 29 & 501 M & 500 K & 265.30 & 2.30 & 5.60 & 0.82 & 0.58 & 8.21 \\
     %DTR~\cite{park2023denoising}+ANT-UW~\cite{go2024addressing} & 458 M & 2,000 K & - & 2.33 & - & - & - \\
     ShortGPT~\cite{men2024shortgpt} & 28$\Rightarrow$19 & 459 M & 100 K & 132.79 & 7.93 & 5.25 & 0.76 & 0.53 & 10.07 \\
     \grayrow TinyDiT-D19 (KD) & 28$\Rightarrow$19 & 459 M & 100 K & 242.29 & 2.90 & 4.63 & 0.84 & 0.54 & 10.07 \\ %e3
     \grayrow TinyDiT-D19 (KD)  & 28$\Rightarrow$19 & 459 M & 500 K & 251.02 & 2.55 & 4.57 & 0.83 & 0.55 & 10.07 \\ 

    \midrule
     DiT-L/2~\cite{peebles2023scalable} & 24 &458 M & 1,000 K  & 196.26 & 3.73 & 4.62 & 0.82 & 0.54 & 9.73  \\
     U-ViT-L~\cite{bao2023all} & 21 & 287 M & 300 K & 221.29 & 3.44 & 6.58 & 0.83 & 0.52 & 13.48 \\
     U-DiT-L~\cite{tian2024u} & 22 & 204 M & 400 K & 246.03 & 3.37 & 4.49 & 0.86 & 0.50 & - \\
     %DTR & & - & 400 K & 156.48 & 8.90 & - &  0.77 & 0.51 & \\
     Diff-Pruning-50\%~\cite{fang2023structural} & 28 & 338 M & 100 K & 186.02 & 3.85 & 4.92 & 0.82 & 0.54 & 10.43 \\
     Diff-Pruning-75\%~\cite{fang2023structural} & 28 & 169 M & 100 K &  83.78 & 14.58 & 6.28 & 0.72 & 0.53 & 13.59   \\
     ShortGPT~\cite{men2024shortgpt} & 28$\Rightarrow$14 & 340 M & 100 K & 66.10 & 22.28 & 6.20 & 0.63 & 0.56 & 13.54 \\ % [0, 1, 2, 3, 6, 15, 17, 18, 20, 21, 22, 25, 26, 27]
     Flux-Lite~\cite{flux1-lite} & 28$\Rightarrow$14 & 340 M & 100 K & 54.54 & 25.92 & 5.98 & 0.62 & 0.55 & 13.54 \\%[12, 14, 15, 17, 18, 19, 20, 21, 22, 23, 24, 25, 26, 27] 
     Sensitivity Analysis~\cite{han2015learning} & 28$\Rightarrow$14 & 340 M & 100 K & 70.36 & 21.15 & 6.22 & 0.63 & \bf 0.57 & 13.54 \\
     Oracle (BK-SDM)~\cite{kim2023bk} & 28$\Rightarrow$14 & 340 M & 100 K & 141.18 & 7.43 & 6.09 & 0.75 & 0.55 & 13.54 \\ % [ 0,  2,  4,  6,  8, 10, 12, 15, 17, 19, 21, 23, 25, 27]
     %WDPruning~\cite{yu2022width} & \\ % [0,1,2,3,4,5,6,7,8,9,10,11,12,13]
    \grayrow TinyDiT-D14 & 28$\Rightarrow$14 & 340 M & 100 K & 151.88 & 5.73 & 4.91 &  0.80 & 0.55 & 13.54 \\
    \grayrow TinyDiT-D14 & 28$\Rightarrow$14 & 340 M & 500 K & 198.85 & 3.92 & 5.69 & 0.78 & 0.58 & 13.54 \\
    \grayrow  TinyDiT-D14 (KD) & 28$\Rightarrow$14 & 340 M & 100 K & 207.27 & 3.73 & 5.04 & 0.81 & 0.54 & 13.54  \\
    %\grayrow  TinyDiT-D14 & 28$\Rightarrow$14 & 340 M & 500 K & 234.54 & 2.90 & 4.80 & 0.82 & 0.55 & 12.76it/s \\
    \grayrow  TinyDiT-D14 (KD) & 28$\Rightarrow$14 & 340 M & 500 K & 234.50 & 2.86 & 4.75 & 0.82 & 0.55 & 13.54  \\  % outputs/Final-D14-DiT-D14-2/checkpoints/0500000.pt

    \midrule
    DiT-B/2~\cite{peebles2023scalable} & 12 & 130 M & 1,000 K  & 119.63 & 10.12 & 5.39 & 0.73 & 0.55 & 28.30  \\
    %DiT-B/2 & 130 M & 400 K & 64.72 & 27.96 &  & 0.57 & 0.52 \\
    U-DiT-B~\cite{tian2024u} & 22 & - & 400 K & 85.15 & 16.64 & 6.33 & 0.64 & 0.63 & - \\
    %DiT-S/2 & 12 & 32 M & 1,000 K  & 48.40 & 32.55  & 7.70 & 0.55 & 0.57  \\
    %DiT-S/2 & 12 & 32 M & 2,000 K  & 59.39 & 26.51 & 7.31 & 0.59 & 0.56 & 63.60  \\
    %\grayrow TinyDiT-D7 & 14$\Rightarrow$7 & 173 M & 500 K & 166.01 & 6.11 & 5.69 & 0.77 & 0.54 \\
    %\grayrow TinyDiT-D7 (KD) & 28$\Rightarrow$7 & 173 M & 500 K & 137.88 & 8.80 & 5.92 & 0.73 & 0.55 & 26.29 \\
    \grayrow TinyDiT-D7 (KD) & 14$\Rightarrow$7 & 173 M & 500 K & 166.91 & 5.87 & 5.43 & 0.78 & 0.53 & 26.81 \\
    %\grayrow TinyDiT-4 & 7$\Rightarrow$4 & 101 M & 500K & 84.14 & 19.89 & 7.83 & 0.6284 & 0.5387 & 43.67 \\
    %\grayrow TinyDiT-2 & 4$\Rightarrow$2 & 53 M  & & & & & & & 79.09 \\

    \bottomrule
    \end{tabular}
    %}
    \caption{Layer pruning results for pre-trained DiT-XL/2. We focus on two settings: fast training with 100K optimization steps and sufficient fine-tuning with 500K steps. Both fine-tuning and Masked Knowledge Distillation (a variant of KD, see Sec.~\ref{sec:kd}) are used for recovery.}
    \label{tab:main_dits}
\end{table*}

\paragraph{Joint Optimization with Recoverability.} With differentiable sampling, we are able to update the underlying probability using gradient descent. The training objective in this work is to maximize the recoverability of sampled masks. We reformulate the objective in Equation~\ref{eqn:objective} by incorporating the learnable distribution:
\begin{equation}
     \min_{\{p(\mask_k)\}} \underbrace{\min_{\Delta\Phi} \;  \mathbb{E}_{x, \{\mask_k\sim p(\mask_k)\}} \left[ \mathcal{L}(x, \Phi+\Delta\Phi, \{\mask_k\}\right]}_{\textit{Recoverability: Post-Fine-Tuning Performance}}, \label{eqn:general_objective}
\end{equation}
where $\{p(\mask_k)\}=\{p(\mask_1), \cdots, p(\mask_K)\}$ refer to the categorical distributions for different local blocks. Based on this formulation, we further investigate how to incorporate the fine-tuning information into the training. We propose a joint optimization of the distribution and a weight update $\Delta \Phi$. Our key idea is to introduce a co-optimized update $\Delta\Phi$ for joint training. A straightforward way to craft the update is to directly optimize the original network. However, the parameter scale in a diffusion transformer is usually huge, and a full optimization may make the training process costly and not that efficient. To this end, we show that Parameter-Efficient Fine-Tuning methods such as LoRA~\cite{hu2022lora} can be a good choice to obtain the required $\Delta\Phi$. For a single linear matrix $\mathbf{W}$ in $\Phi$, we simulate the fine-tuned weights as:
\begin{equation}
    \mathbf{W}_{\text{fine-tuned}} = \mathbf{W} + \alpha \Delta \mathbf{W} = \mathbf{W} + \alpha \mathbf{B} \mathbf{A},
\end{equation}
where $\alpha$ is a scalar hyperparameter that scales the contribution of $\Delta \mathbf{W}$. Using LoRA significantly reduces the number of parameters, facilitating efficient exploration of different pruning decisions. As shown in Figure~\ref{fig:lora}, we leverage the sampled binary mask value $m_i$ as the gate and forward the network with Equation~\ref{eqn:forward}, which suppresses the layer outputs if the sampled mask is 0 for the current layer. In addition, the previously mentioned STE will still provide non-zero gradients to the pruned layer, allowing it to be further updated. This is helpful in practice, since some layers might not be competitive at the beginning, but may emerge as competitive candidates with sufficient fine-tuning. 

\paragraph{Pruning Decision.} After training, we retain those local structures with the highest probability and discard the additional update $\Delta\Phi$. Then, standard fine-tuning techniques can be applied for recovery.  %The whole pipeline is summarized in Algorithm~\ref{alg:alg}.

\section{Experiments}

\subsection{Experimental Settings}

Our experiments were mainly conducted on Diffusion Transformers~\cite{peebles2023scalable} for class-conditional image generation on ImageNet 256 $\times$ 256~\cite{deng2009imagenet}. For evaluation, we follow~\cite{dhariwal2021diffusion,peebles2023scalable} and report the Fréchet inception distance (FID), Sliding Fréchet Inception Distance (sFID), Inception Scores (IS), Precision and Recall using the official reference images~\cite{dhariwal2021diffusion}. Additionally, we also extend our methods to other models, including MARs~\cite{li2024autoregressive} and SiTs~\cite{ma2024sit}. Experimental details can be found in the following sections and appendix.%\textbf{DiT} section (see~\ref{par:exp_dits}) and the \textbf{MAR \& SiT} section (see~\ref{par:exp_others}).

\begin{figure}[t]
    \centering
    \includegraphics[width=\linewidth]{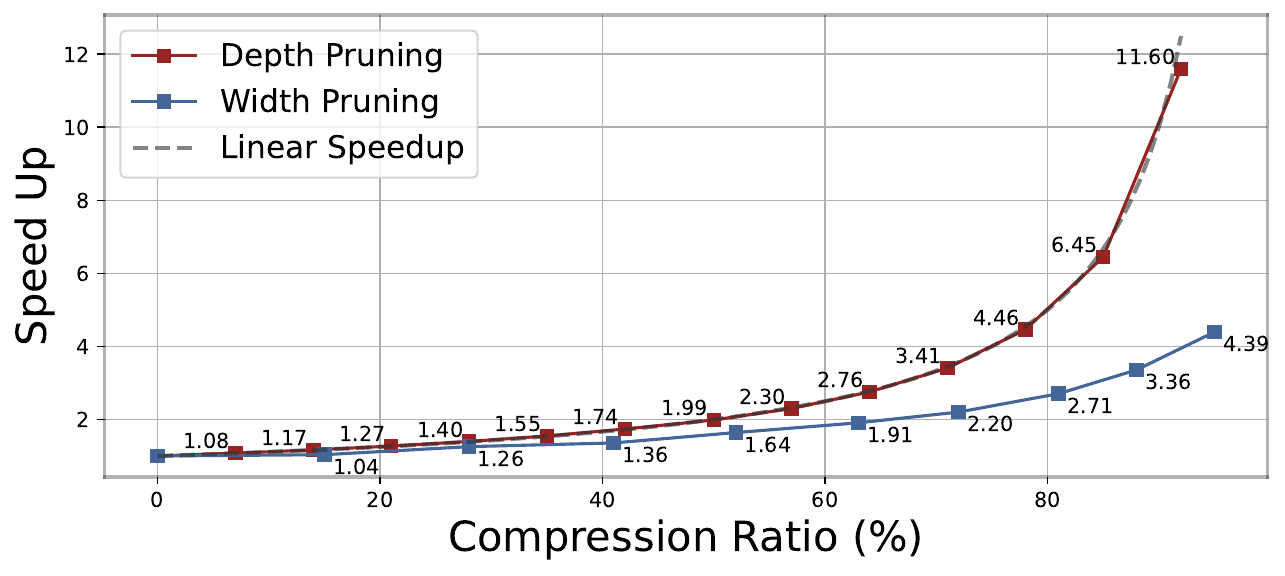}
    \caption{Depth pruning closely aligns with the theoretical linear speed-up relative to the compression ratio.}
    \label{fig:speed}
\end{figure}

\subsection{Results on Diffusion Transformers} 

\paragraph{DiT.}

This work focuses on the compression of DiTs~\cite{peebles2023scalable}. We consider two primary strategies as baselines: the first involves using manually crafted patterns to eliminate layers. For instance, BK-SDM~\cite{kim2023bk} employs heuristic assumptions to determine the significance of specific layers, such as the initial or final layers. The second strategy is based on systematically designed criteria to evaluate layer importance, such as analyzing the similarity between block inputs and outputs to determine redundancy~\cite{men2024shortgpt,flux1-lite}; this approach typically aims to minimize performance degradation after pruning. Table~\ref{tab:main_dits} presents representatives from both strategies, including ShortGPT~\cite{men2024shortgpt}, Flux-Lite~\cite{flux1-lite},  Diff-Pruning~\cite{fang2023structural}, Sensitivity Analysis~\cite{han2015learning} and BK-SDM~\cite{kim2023bk}, which serve as baselines for comparison. Additionally, we evaluate our method against innovative architectural designs, such as UViT~\cite{bao2023all}, U-DiT~\cite{tian2024u}, and DTR~\cite{park2023denoising}, which have demonstrated improved training efficiency over conventional DiTs.

Table~\ref{tab:main_dits} presents our findings on compressing a pre-trained DiT-XL/2~\cite{peebles2023scalable}. This model contains 28 transformer layers structured with alternating Attention and MLP layers. The proposed method seeks to identify shallow transformers with $\{7, 14, 19\}$ sub-layers from these 28 layers, to maximize the post-fine-tuning performance. With only 7\% of the original training cost (500K steps compared to 7M steps), TinyDiT achieves competitive performance relative to both pruning-based methods and novel architectures. For instance, a DiT-L model trained from scratch for 1M steps achieves an FID score of 3.73 with 458M parameters. In contrast, the compressed TinyDiT-D14 model,  with only 340M parameters and a faster sampling speed (13.54 it/s vs. 9.73 it/s), yields a significantly improved FID of 2.86. On parallel devices like GPUs, the primary bottleneck in transformers arises from sequential operations within each layer, which becomes more pronounced as the number of layers increases. Depth pruning mitigates this bottleneck by removing entire transformer layers, thereby reducing computational depth and optimizing the workload. By comparison, width pruning only reduces the number of neurons within each layer, limiting its speed-up potential. As shown in Figure~\ref{fig:speed}, depth pruning closely matches the theoretical linear speed-up as the compression ratio increases, outperforming width pruning methods such as Diff-Pruning~\cite{fang2023structural}.

\begin{table}[t]
    \centering
    \small
    \resizebox{\linewidth}{!}{
    \begin{tabular}{l|c c c c c}
    \toprule
      Method & Depth & Params & Epochs & FID & IS \\
        \midrule
       MAR-Large & 32 & 479 M & 400 & 1.78 & 296.0 \\
         MAR-Base & 24 & 208 M & 400 & 2.31 & 281.7 \\
        %& TinyMAR-D16 Finetune & 16 & 277 M & 40  & 3.01 & 235.7 \\
        %TinyMAR-D16 & 16 & 277 M & 40  & 2.35 & 283.6 \\
        TinyMAR-D16 & 32$\Rightarrow$16 & 277 M & 40  & 2.28 & 283.4 \\ 
        \midrule
        SiT-XL/2 & 28 & 675 M & 1,400 & 2.06 & 277.5 \\
        % & TinySiT-D14 Finetune & 14 & 340 M & 100  & 3.16 & 218.8 \\
        %& TinySiT-D14 KD & 14 & 340 M & 100  &  & \\
        TinySiT-D14 & 28$\Rightarrow$14 & 340 M & 100 & 3.02 & 220.1 \\ % 220.08 & 3.02 & 4.70 & 0.8081 & 0.5708
    \bottomrule
    \end{tabular}
    }
    \caption{Depth pruning results on MARs~\cite{li2024autoregressive} and SiTs~\cite{ma2024sit}.}
    \label{tab:mar_sit}
\end{table}

\paragraph{MAR \& SiT.} Masked Autoregressive (MAR)~\cite{li2024autoregressive} models employ a diffusion loss-based autoregressive framework in a continuous-valued space, achieving high-quality image generation without the need for discrete tokenization. The MAR-Large model, with 32 transformer blocks, serves as the baseline for comparison. Applying our pruning method, we reduced MAR to a 16-block variant, TinyMAR-D16, achieving an FID of 2.28 and surpassing the performance of the 24-block MAR-Base model with only 10\% of the original training cost (40 epochs vs. 400 epochs). Our approach also generalizes to Scalable Interpolant Transformers (SiT)~\cite{ma2024sit}, an extension of the DiT architecture that employs a flow-based interpolant framework to bridge data and noise distributions. The SiT-XL/2 model, comprising 28 transformer blocks, was pruned by 50\%, creating the TinySiT-D14 model. This pruned model retains competitive performance at only 7\% of the original training cost (100 epochs vs. 1400 epochs). As shown in Table~\ref{tab:mar_sit}, these results demonstrate that our pruning method is adaptable across different diffusion transformer variants, effectively reducing the model size and training time while maintaining strong performance.

\subsection{Analytical Experiments}

\begin{figure}[t]
    \centering
    \includegraphics[width=\linewidth]{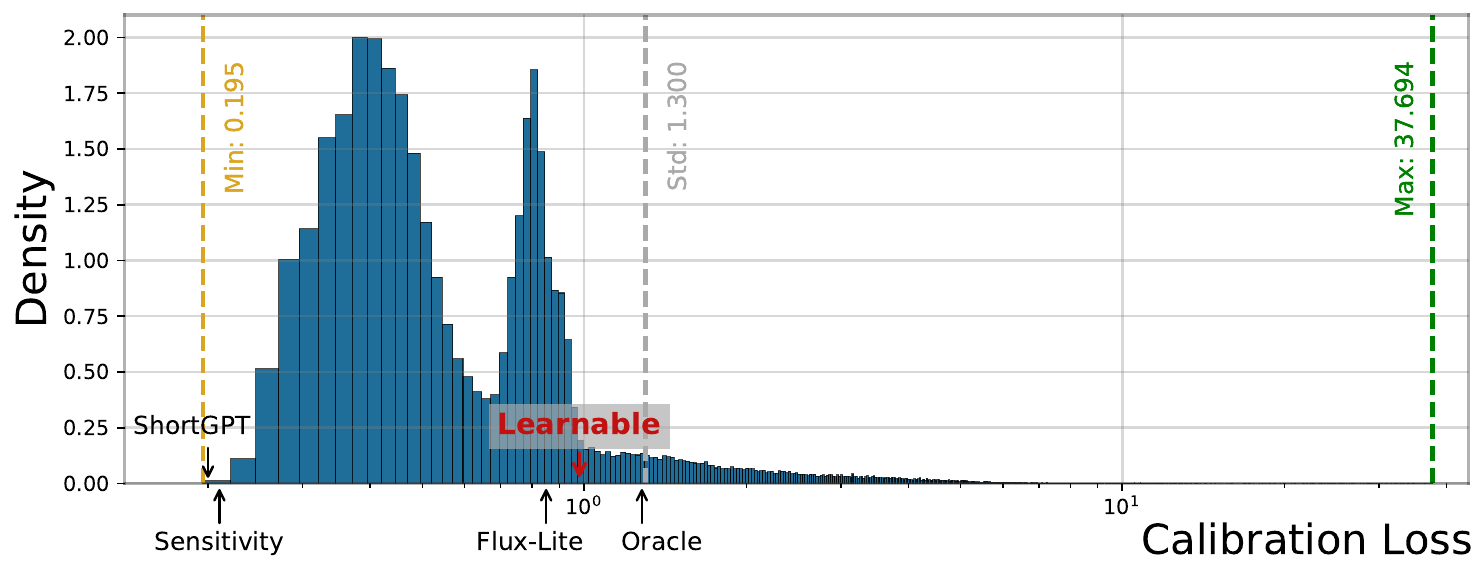}
    \vspace{-0.8cm}
    \caption{Distribution of calibration loss through random sampling of candidate models. The proposed learnable method achieves the best post-fine-tuning FID yet has a relatively high initial loss compared to other baselines.}
    \label{fig:distribution}
\end{figure}

\begin{table}[t!]
    \centering
   % \small
    \resizebox{\linewidth}{!}{
    \begin{tabular}{l|c c c c c}
    \toprule
        \bf Strategy & \bf Loss & \bf IS & \bf FID & \bf Prec. & \bf Recall \\
        \midrule
        Max. Loss & 37.69 & NaN & NaN & NaN & NaN \\%[2, 6, 7, 8, 9, 11, 12, 15, 16, 18, 20, 21, 22, 27] \\
        %Med. Loss & 0.31 & 120.04 & 9.59 & 5.72 & 0.74 & 0.55 \\ %&  [1, 2, 4, 6, 7, 10, 12, 17, 18, 19, 23, 24, 25, 27] \\
        Med. Loss & 0.99 & 149.51 & 6.45 & 0.78 & 0.53 \\% [0, 1, 2, 3, 5, 6, 8, 9, 11, 12, 13, 14, 20, 25]
        Min. Loss & \bf 0.20 & 73.10 & 20.69 & 0.63 & 0.58 \\%[0, 1, 2, 3, 4, 6, 8, 18, 19, 21, 22, 25, 26, 27]  \\ & 6.39
        Sensitivity & 0.21 & 70.36 & 21.15  & 0.63 & \bf 0.57 \\%[0, 1,  2, 3, 4, 6, 15, 18, 19, 22, 23, 24, 26, 27] \\ & 6.22
        ShortGPT~\cite{men2024shortgpt} & 0.20 & 66.10 & 22.28  & 0.63 & 0.56 \\%[0, 1, 2, 3, 6, 15, 17, 18, 20, 21, 22, 25, 26, 27] \\ & 6.20
        Flux-Lite~\cite{flux1-lite} & 0.85 & 54.54 & 25.92 & 0.62 & 0.55 \\%[12, 14, 15, 17, 18, 19, 20, 21, 22, 23, 24, 25, 26, 27] \\ & NaN 
        %\midrule
        Oracle (BK-SDM) & 1.28 & 141.18 & 7.43  & 0.75 & 0.55 \\%[0, 2, 4, 6, 8, 10, 12, 14, 16, 18, 20, 22, 24, 27] \\ & 6.09
        %Sensitivity  & 1.65 & 144.92 & 6.89  & 0.76 & 0.54 & 3 \\%[1, 2, 4, 6, 9, 10, 12, 15, 17, 18, 21, 22, 24, 26] \\& 5.65
        \grayrow Learnable & 0.98 & \bf 151.88 & \bf 5.73 & \bf 0.80 & 0.55 \\ %[0, 2, 4, 6, 8, 10, 13, 14, 16, 18, 20, 23, 24, 26] \\ & \bf 4.91 
    \bottomrule
    \end{tabular}
    }
    \caption{Directly minimizing the calibration loss may lead to non-optimal solutions. All pruned models are fine-tuned \emph{without} knowledge distillation (KD) for 100K steps. We evaluate the following baselines: (1) Loss – We randomly prune a DiT-XL model to generate 100,000 models and select models with different calibration losses for fine-tuning; (2) Metric-based Methods – such as Sensitivity Analysis and ShortGPT; (3) Oracle – We retain the first and last layers while uniformly pruning the intermediate layers following \cite{kim2023bk}; (4) Learnable – The proposed learnable method.}
    \label{tab:loss_value}
\end{table}

\paragraph{Is Calibration Loss the Primary Determinant?}  
An essential question in depth pruning is how to identify redundant layers in pre-trained diffusion transformers. A common approach involves minimizing the calibration loss, based on the assumption that a model with lower calibration loss after pruning will exhibit superior performance. However, we demonstrate in this section that this hypothesis may not hold for diffusion transformers. We begin by examining the solution space through random depth pruning at a 50\% ratio, generating 100,000 candidate models with calibration losses ranging from 0.195 to 37.694 (see Figure~\ref{fig:distribution}). From these candidates, we select models with the highest and lowest calibration losses for fine-tuning. Notably, both models result in unfavorable outcomes, such as unstable training (NaN) or suboptimal FID scores (20.69), as shown in Table~\ref{tab:loss_value}. Additionally, we conduct a sensitivity analysis~\cite{han2015learning}, a commonly used technique to identify crucial layers by measuring loss disturbance upon layer removal, which produces a model with a low calibration loss of 0.21. However, this model’s FID score is similar to that of the model with the lowest calibration loss. Approaches like ShortGPT~\cite{men2024shortgpt} and a recent approach for compressing the Flux model~\cite{flux1-lite}, which estimate similarity or minimize mean squared error (MSE) between input and output states, reveal a similar trend. In contrast, methods with moderate calibration losses, such as Oracle (often considered less competitive) and one of the randomly pruned models, achieve FID scores of 7.43 and 6.45, respectively, demonstrating significantly better performance than models with minimal calibration loss. These findings suggest that, while calibration loss may influence post-fine-tuning performance to some extent, it is not the primary determinant for diffusion transformers. Instead, the model's capacity for performance recovery during fine-tuning, termed ``recoverability,'' appears to be more critical. Notably, assessing recoverability using traditional metrics is challenging, as it requires a learning process across the entire dataset. This observation also explains why the proposed method achieves superior results (5.73) compared to baseline methods.

\begin{table}[t]
    \centering
    \resizebox{\linewidth}{!}{
    \begin{tabular}{l l| c c c c c}
    \toprule
        \bf Pattern  & $\mathbf{\Delta}$\bf{W} & \bf IS $\uparrow$ & \bf FID $\downarrow$ & \bf sFID $\downarrow$ & \bf Prec. $\uparrow$ & \bf Recall $\uparrow$ \\
        %\midrule
        %\midrule
        %1:2 & 118.71 & 13.56 & 20.37 & 0.7162 & 0.6203 \\%[1, 2, 4, 6, 8, 10, 12, 14, 16, 18, 20, 23, 24, 26]
        %2:4 & 110.73 & 15.29 & 21.28 & 0.7032 & 0.6303  \\%[1, 2, 4, 6, 8, 9, 12, 13, 16, 17, 20, 23, 24, 25]
        %7:14 & 73.99 & 24.45 & 20.98 & 0.6114 & 0.6327\\%[0, 1, 2, 4, 6, 7, 8, 16, 17, 18, 19, 22, 23, 24]
        
        %\multicolumn{7}{c}{\textbf{Prune by learning (train)}} \\
    
        \midrule
        1:2 & LoRA  & 54.75 & 33.39 & 29.56 & 0.56 & 0.62 \\%[0, 2, 4, 6, 8, 10, 13, 14, 16, 18, 20, 23, 24, 26]
        %Lora - 1:2 Prev SOTA  & 47.68 & 37.81 & 28.65 & 0.5351 & 0.5998 \\%[1,  2,  4,  6,  9, 11, 13, 15, 16, 19, 21, 23, 24, 26]
        %2:4 & LoRA & 42.25 & 41.87 & 30.52 & 0.5185 & 0.5949 \\%[1, 2, 4, 6, 8, 9, 13, 15, 18, 19, 20, 23, 24, 26]
        2:4 & LoRA & 53.07 & 34.21 & 27.61 & 0.55 & 0.63 \\ % [0, 2, 6, 7, 8, 9, 12, 13, 18, 19, 20, 23, 24, 25]
        
        %Lora - 2:4 epoch-14 & 50.86 & 36.64 & 31.20 & 0.5554 & 0.5981 \\%[1, 2, 6, 7, 8, 9, 12, 13, 16, 17, 20, 23, 24, 25]
        7:14 & LoRA & 34.97 & 49.41 & 28.48 & 0.46 & 0.56 \\%[0, 1, 2, 4, 5, 6, 7, 15, 18, 19, 23, 24, 25, 26]
        %7:14 & LoRA rank-32 & 34.11 & 50.77 & 30.23 & 0.4560 & 0.5691 \\ %[[0, 1, 2, 4, 5, 6, 7, 16, 17, 18, 23, 24, 25, 26]
        %7:14 & LoRA rank-128 & 46.02 & 38.94 & 29.79 & 0.5270 & 0.6168\\ %[1, 2, 6, 7, 8, 9, 12, 15, 18, 19, 23, 24, 25, 26]
        \midrule
        1:2 & Full  & 53.11 & 35.77 & 32.68 & 0.54 & 0.61 \\ % [0, 2, 4, 6, 8, 10, 12, 14, 16, 18, 20, 23, 24, 26]
        2:4 & Full & 53.63 & 34.41 & 29.93 & 0.55 & 0.62 \\ % [0, 1, 6, 7, 8, 9, 12, 13, 16, 19, 20, 23, 24, 26]
        7:14 & Full & 45.03 & 38.76 & 31.31 & 0.52 & 0.62 \\ % [1, 2, 6, 7, 8, 9, 13, 15, 16, 19, 20, 21, 23, 24]

        \midrule
        1:2 & Frozen & 45.08 & 39.56 & 31.13 & 0.52 & 0.60 \\ % [1, 2, 4, 6, 9, 11, 13, 15, 16, 19, 20, 23, 24, 26]
        2:4 & Frozen & 48.09 & 37.82 & 31.91 & 0.53 & 0.62 \\ % [1, 2, 4, 6, 8, 9, 12, 13, 16, 19, 21, 23, 24, 26]
        7:14 & Frozen & 34.09 & 49.75 & 31.06 & 0.46 & 0.56 \\ % [0, 1, 2, 3, 5, 6, 9, 14, 16, 19, 21, 23, 24, 26]
        %\midrule

    \bottomrule
    \end{tabular}
    }
    \caption{Performance comparison of TinyDiT-D14 models compressed using various pruning schemes and recoverability estimation strategies. All models are fine-tuned for 10,000 steps, and FID scores are computed on 10,000 sampled images with 64 timesteps.}
    \label{tab:block}
\end{table}

\paragraph{Learnable Modeling of Recoverability.} To overcome the limitations of traditional metric-based methods, this study introduces a learnable approach to jointly optimize pruning and model recoverability. Table~\ref{tab:loss_value} illustrates different configurations of the learnable method, including the local pruning scheme and  update strategies for recoverability estimation. For a 28-layer DiT-XL/2 with a fixed 50\% layer pruning rate, we examine three splitting schemes: 1:2, 2:4, and 7:14. In the 1:2 scheme, for example, every two transformer layers form a local block, with one layer pruned. Larger blocks introduce greater diversity but significantly expand the search space. For instance, the 7:14 scheme divides the model into two segments, each retaining 7 layers, resulting in $\binom{14}{7} \times 2 = 6{,}864$ possible solutions. Conversely, smaller blocks significantly reduce optimization difficulty and offer greater flexibility. When the distribution of one block converges, the learning on other blocks can still progress. As shown in Table~\ref{tab:loss_value}, the 1:2 configuration achieves the optimal performance after 10K fine-tuning iterations. Additionally, our empirical findings underscore the effectiveness of recoverability estimation using LoRA or full fine-tuning. Both methods yield positive post-fine-tuning outcomes, with LoRA achieving superior results (FID = 33.39) compared to full fine-tuning (FID = 35.77) under the 1:2 scheme, as LoRA has fewer trainable parameters (0.9\% relative to full parameter training) and can adapt more efficiently to the randomness of sampling.

\begin{figure}[t]
    \centering
    \includegraphics[width=\linewidth]{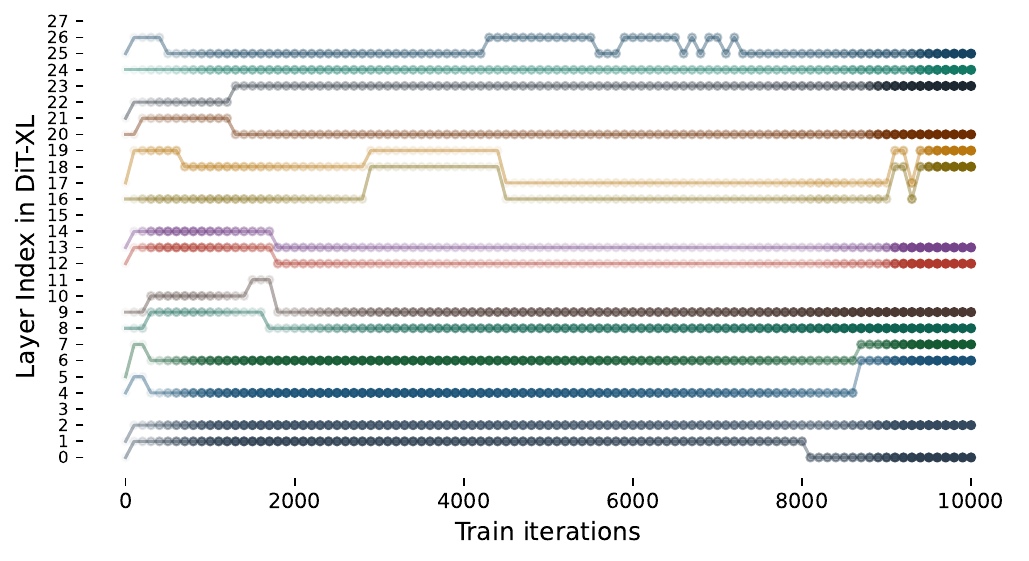}
    \caption{Visualization of the 2:4 decisions in the learnable pruning, with the confidence level of each decision highlighted through varying degrees of transparency. More visualization results for 1:2 and 7:14 schemes are available in the appendix.}
    \label{fig:learnable_pruning_indices}
\end{figure}

\begin{figure*}[t]
    \centering
    \includegraphics[width=\linewidth]{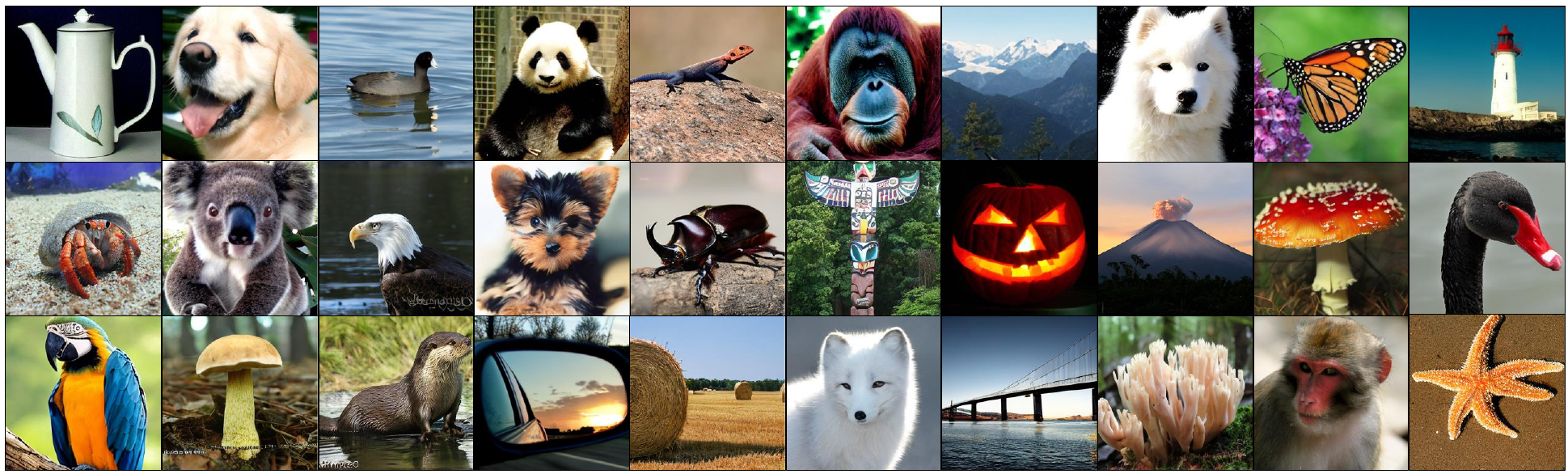}
    \caption{Images generated by TinyDiT-D14 on ImageNet 224$\times$224, pruned and distilled from a DiT-XL/2.}
    \label{fig:vis}
\end{figure*}

\paragraph{Visualization of Learnable Decisions.} To gain deeper insights into the role of the learnable method in pruning, we visualize the learning process in Figure~\ref{fig:learnable_pruning_indices}. From bottom to top, the i-th curve represents the i-th layer of the pruned model, displaying its layer index in the original DiT-XL/2. This visualization illustrates the dynamics of pruning decisions over training iterations, where the transparency of each data point indicates the probability of being sampled. The learnable method shows its capacity to explore and handle various layer combinations. Pruning decisions for certain layers, such as the 7-th and 8-th in the compressed model, are determined quickly and remain stable throughout the process. In contrast, other layers, like the 0-th layer, require additional fine-tuning to estimate their recoverability. Notably, some decisions may change in the later stages once these layers have been sufficiently optimized. The training process ultimately concludes with high sampling probabilities, suggesting a converged learning process with distributions approaching a one-hot configuration. After training, we select the layers with the highest probabilities for subsequent fine-tuning.

\subsection{Knowledge Distillation for Recovery} \label{sec:kd}

In this work, we also explore Knowledge Distillation (KD) as an enhanced fine-tuning method. As demonstrated in Table~\ref{tab:KD}, we apply the vanilla knowledge distillation approach proposed by Hinton~\cite{hinton2015distilling} to fine-tune a TinyDiT-D14, using the outputs of the pre-trained DiT-XL/2 as a teacher model for supervision. We employ a Mean Square Error (MSE) loss to align the outputs between the shallow student model and the deeper teacher model, which effectively reduces the FID at 100K steps from 5.79 to 4.66.

\begin{figure}[t]
    \centering
    \begin{subfigure}{0.49\linewidth}
        \includegraphics[width=\linewidth]{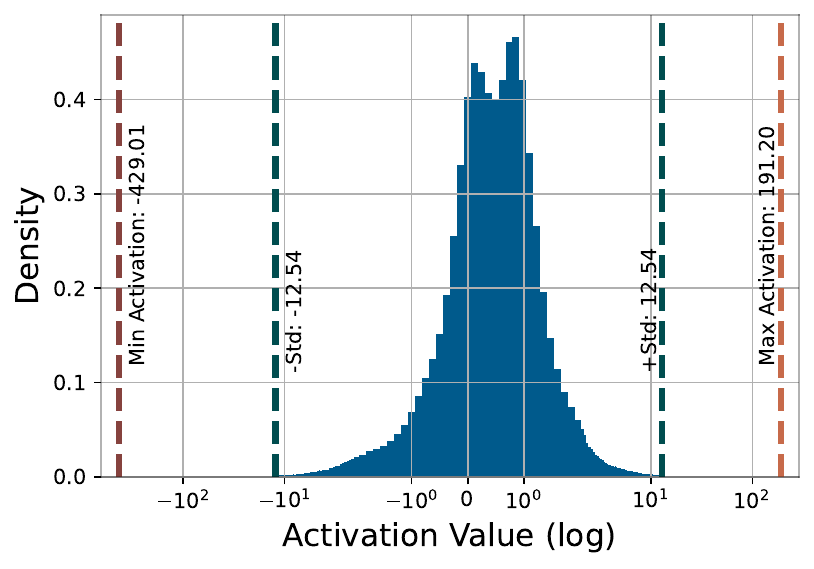}
        \caption{DiT-XL/2 (Teacher)}
    \end{subfigure}
    \begin{subfigure}{0.49\linewidth}
        \includegraphics[width=\linewidth]{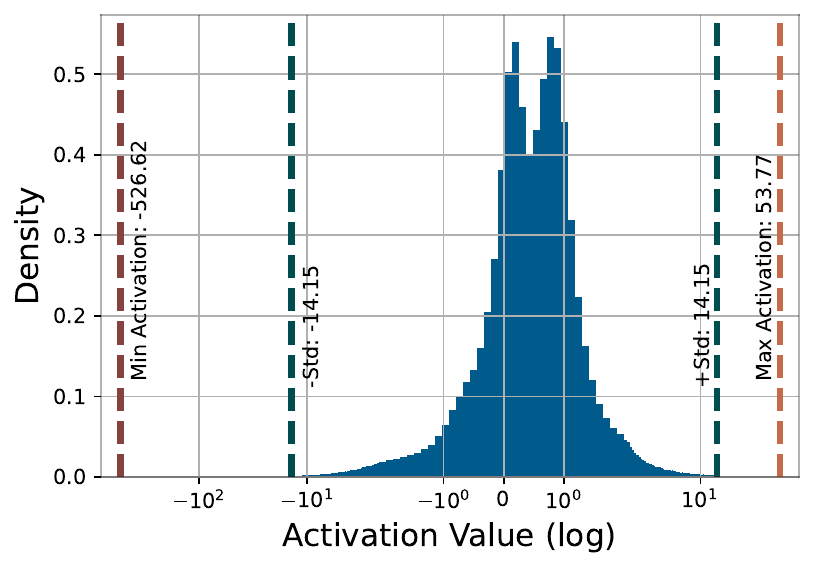}
        \caption{TinyDiT-D14 (Student)}
    \end{subfigure}
    
    \caption{Visualization of massive activations~\cite{sun2024massive} in DiTs. Both teacher and student models display large activation values in their hidden states. Directly distilling these massive activations may result in excessively large losses and unstable training.}
    \label{fig:activation}
\end{figure}

\begin{table}[t]
    \centering
    \small
    %\resizebox{\linewidth}{!}{
    \begin{tabular}{l | c c }
    \toprule
        \bf fine-tuning Strategy & \bf Init. Distill. Loss & \bf FID @ 100K \\
        \midrule
        fine-tuning & - & 5.79  \\
        Logits KD & - & 4.66 \\
        \midrule
        RepKD & 2840.1 & NaN \\
        Masked KD ($0.1\sigma$) & 15.4 & NaN   \\
        %Masked KD ($1\sigma$) &  \\
        Masked KD ($2\sigma$) & 387.1 & 3.73 \\
        Masked KD ($4\sigma$) & 391.4 & 3.75 \\
    \bottomrule
    \end{tabular}
    %}
    \caption{Evaluation of different fine-tuning strategies for recovery. Masked RepKD ignores those massive activations ($|x|>k\sigma_x$) in both teacher and student, which enables effective knowledge transfer between diffusion transformers.}
    \label{tab:KD}
\end{table}

\paragraph{Masked Knowledge Distillation.} Additionally, we evaluate representation distillation (RepKD) \cite{romero2014fitnets, kim2023bk} to transfer hidden states from the teacher to the student. It is important to note that depth pruning does not alter the hidden dimension of diffusion transformers, allowing for direct alignment of intermediate hidden states. For practical implementation, we use the block defined in Section~\ref{sec:local_structure} as the basic unit, ensuring that the pruned local structure in the pruned DiT aligns with the output of the original structure in the teacher model. However, we encountered significant training difficulties with this straightforward RepKD approach due to massive activations in the hidden states, where both teacher and student models occasionally exhibit large activation values, as shown in Figure~\ref{fig:activation}. Directly distilling these extreme activations can result in excessively high loss values, impairing the performance of the student model. This issue has also been observed in other transformer-based generative models, such as certain LLMs~\cite{sun2024massive}. To address this, we propose a Masked RepKD variant that selectively excludes these massive activations during knowledge transfer. We employ a simple thresholding method, $|x-\mu_x|<k\sigma_x$, which ignores the loss associated with these extreme activations. As shown in Table~\ref{tab:KD}, the Masked RepKD approach with moderate thresholds of $2\sigma$ and $4\sigma$ achieves satisfactory results, demonstrating the robustness of our method. 

\paragraph{Generated Images.} In Figure~\ref{fig:vis}, We visualize the generated images of the learned TinyDiT-D14, distilled from an off-the-shelf DiT-XL/2 model. More visualization results for SiTs and MARs can be found in the appendix. 
%The pruned model is 2$\times$ faster than the original one. With only 500K training steps, the shallow model is able to generate plausible images in the size of $256\times 256$. 

\section{Conclusions}

This work introduces TinyFusion, a learnable method for accelerating diffusion transformers by removing redundant layers. It models the recoverability of pruned models as an optimizable objective and incorporates differentiable sampling for end-to-end training. Our method generalizes to various architectures like DiTs, MARs and SiTs.

{\small
\bibliographystyle{ieee_fullname}
\bibliography{citation}
}

\clearpage

\setcounter{page}{1}
\maketitlesupplementary

% \section{Rationale}
% \label{sec:rationale}
% % 
% Having the supplementary compiled together with the main paper means that:
% % 
% \begin{itemize}
% \item The supplementary can back-reference sections of the main paper, for example, we can refer to \cref{sec:intro};
% \item The main paper can forward reference sub-sections within the supplementary explicitly (e.g. referring to a particular experiment); 
% \item When submitted to arXiv, the supplementary will already included at the end of the paper.
% \end{itemize}
% % 
% To split the supplementary pages from the main paper, you can use \href{https://support.apple.com/en-ca/guide/preview/prvw11793/mac#:~:text=Delete%20a%20page%20from%20a,or%20choose%20Edit%20%3E%20Delete).}{Preview (on macOS)}, \href{https://www.adobe.com/acrobat/how-to/delete-pages-from-pdf.html#:~:text=Choose%20%E2%80%9CTools%E2%80%9D%20%3E%20%E2%80%9COrganize,or%20pages%20from%20the%20file.}{Adobe Acrobat} (on all OSs), as well as \href{https://superuser.com/questions/517986/is-it-possible-to-delete-some-pages-of-a-pdf-document}{command line tools}.

\section{Experimental Details}

\paragraph{Models.} Our experiments evaluate the effectiveness of three models: DiT-XL, MAR-Large, and SiT-XL. Diffusion Transformers (DiTs), inspired by Vision Transformer (ViT) principles, process spatial inputs as sequences of patches. The DiT-XL model features 28 transformer layers, a hidden size of 1152, 16 attention heads, and a 2 $\times$ 2 patch size. It employs adaptive layer normalization (AdaLN) to improve training stability, comprising 675 million parameters and trained for 1400 epochs. Masked Autoregressive models (MARs) are diffusion transformer variants tailored for autoregressive image generation. They utilize a continuous-valued diffusion loss framework to generate high-quality outputs without discrete tokenization. The MAR-Large model includes 32 transformer layers, a hidden size of 1024, 16 attention heads, and bidirectional attention. Like DiT, it incorporates AdaLN for stable training and effective token modeling, with 479 million parameters trained over 400 epochs. Finally, Scalable Interpolant Transformers (SiTs) extend the DiT framework by introducing a flow-based interpolant methodology, enabling more flexible bridging between data and noise distributions. While architecturally identical to DiT-XL, the SiT-XL model leverages this interpolant approach to facilitate modular experimentation with interpolant selection and sampling dynamics.

\paragraph{Datasets.} We prepared the ImageNet 256 $\times$ 256 dataset by applying center cropping and adaptive resizing to maintain the original aspect ratio and minimize distortion. The images were then normalized to a mean of 0.5 and a standard deviation of 0.5. To augment the dataset, we applied random horizontal flipping with a probability of 0.5. To accelerate training without using Variational Autoencoder (VAE), we pre-extracted features from the images using a pre-trained VAE. The images were mapped to their latent representations, normalized, and the resulting feature arrays were saved for direct use during training.

\begin{figure}[t]
    \centering
    \includegraphics[width=\linewidth]{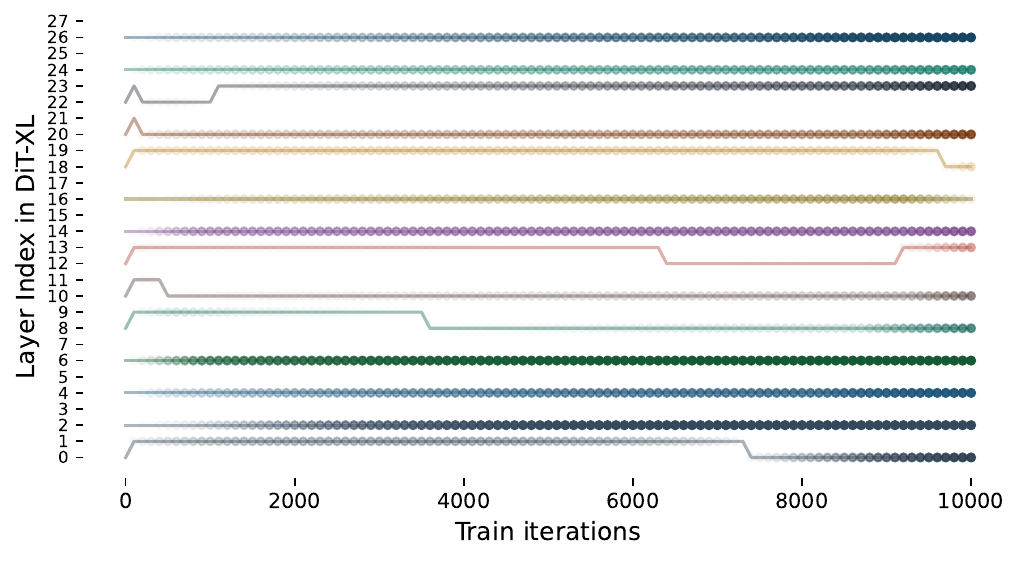}
    \caption{1:2 Pruning Decisions}
    \label{fig:decision-1-2}

    \includegraphics[width=\linewidth]{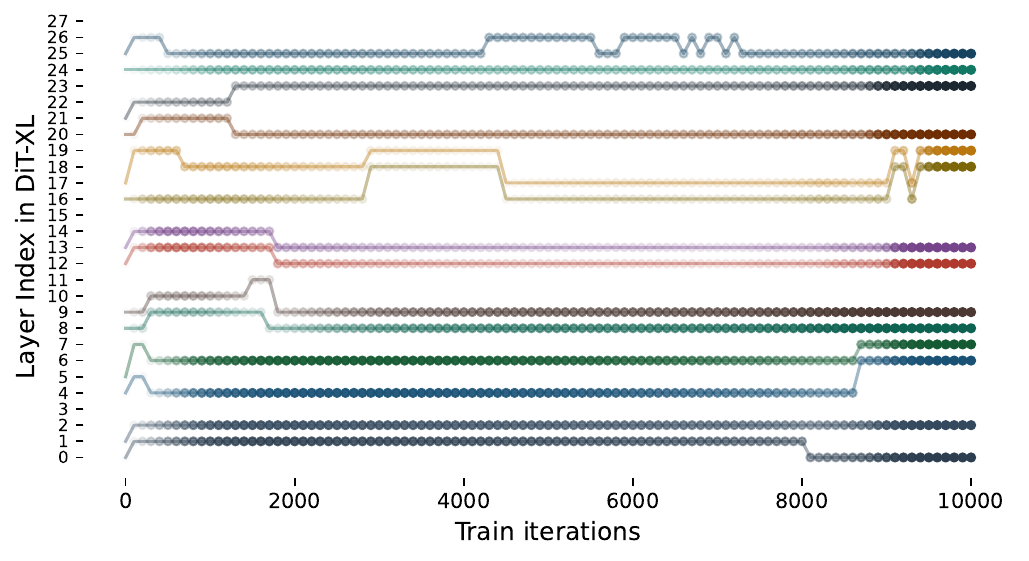}
    \caption{2:4 Pruning Decisions}
    \label{fig:decision-2-4}
    
    \includegraphics[width=\linewidth]{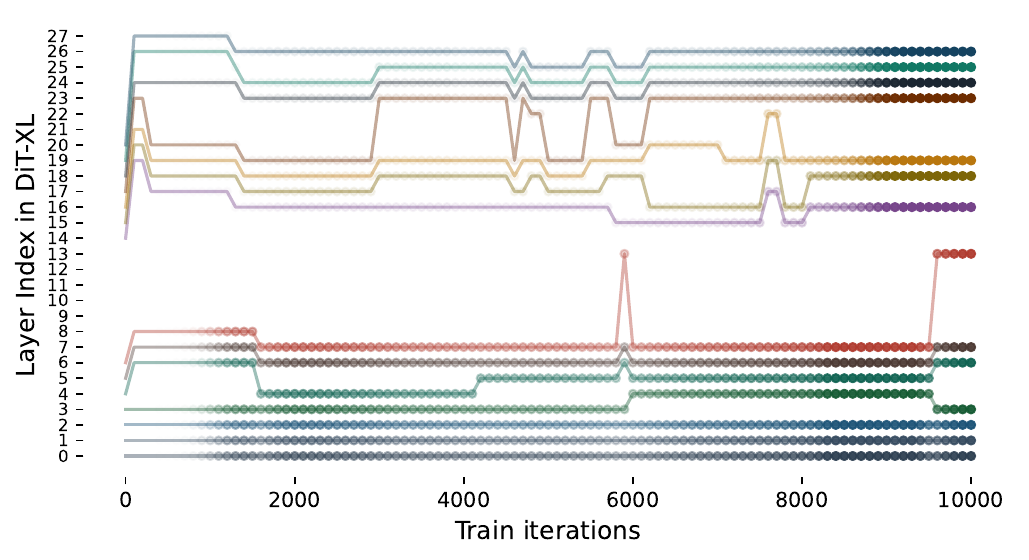}
    \caption{7:14 Pruning Decisions}
    \label{fig:decision-7-14}
\end{figure}

\begin{figure*}
    \centering
    \includegraphics[width=\linewidth]{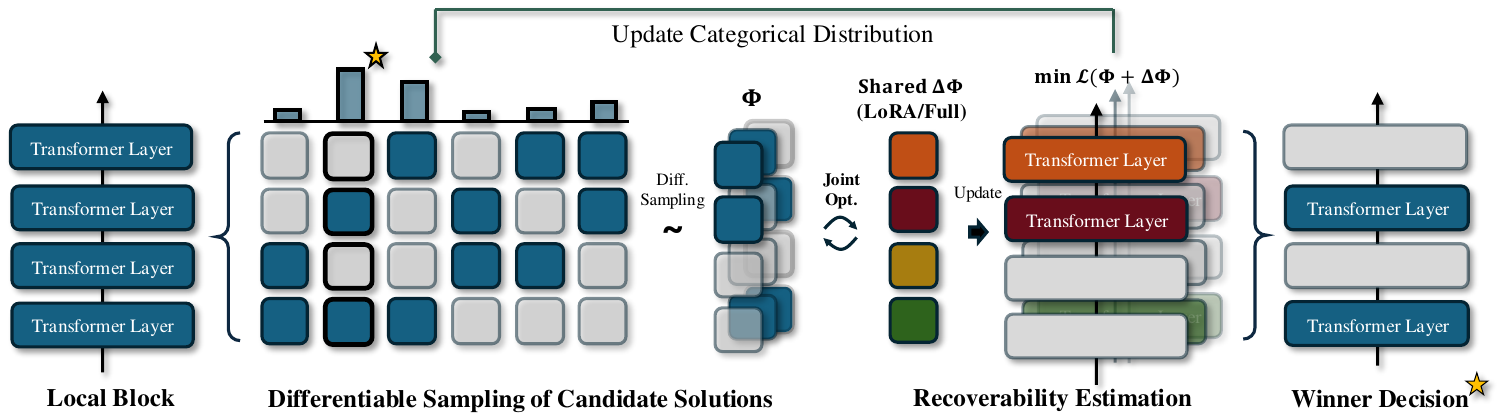}
    \caption{Learnable depth pruning on a local block}
    \label{fig:training_loop}
\end{figure*}

\begin{figure}
    \centering
    \includegraphics[width=\linewidth]{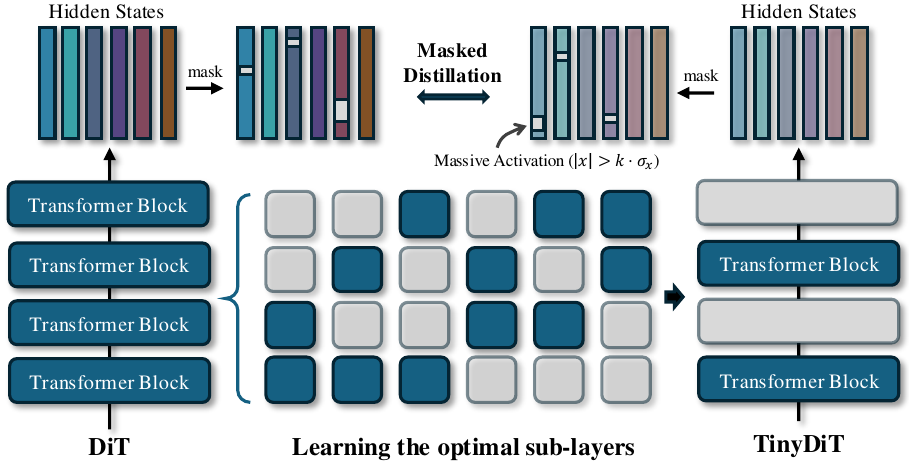}
    \caption{Masked knowledge distillation with 2:4 blocks.}
    \label{fig:kd}
\end{figure}

\paragraph{Training Details} The training process began with obtaining pruned models using the proposed learnable pruning method as illustrated in Figure~\ref{fig:training_loop}. Pruning decisions were made by a joint optimization of pruning and weight updates through LoRA with a block size. In practice, the block size is 2 for simplicity and the models were trained for 100 epochs, except for MAR, which was trained for 40 epochs.  To enhance post-pruning performance, the Masked Knowledge Distillation (RepKD) method was employed during the recovery phase to transfer knowledge from teacher models to pruned student models. The RepKD approach aligns the output predictions and intermediate hidden states of the pruned and teacher models, with further details provided in the following section. Additionally, as Exponential Moving Averages (EMA) are updated and used during image generation, an excessively small learning rate can weaken EMA’s effect, leading to suboptimal outcomes. To address this, a progressive learning rate scheduler was implemented to gradually halve the learning rate throughout training. The details of each hyperparameter are provided in Table~\ref{tab:appendix_hyperparam}.

\begin{table*}[h]
    \centering
    \small
    %\resizebox{\linewidth}{!}{
    \begin{tabular}{l c c c c c c c c c }
    \toprule
        \bf Model & \bf Optimizer & \bf Cosine Sched. & Teacher &\bf $\alpha_{\text{KD}}$ & \bf $\alpha_{\text{GT}}$ & \bf $\beta$ & Grad. Clip & Pruning Configs \\
    \midrule
       DiT-D19 & AdamW(lr=2e-4, wd=0.0) & $\eta_{\text{min}}=1\text{e-4}$ & DiT-XL & 0.9 & 0.1 & 1e-2 $\rightarrow$ 0 & 1.0 & LoRA-1:2 \\
       DiT-D14 & AdamW(lr=2e-4, wd=0.0 & $\eta_{\text{min}}=1\text{e-4}$ & DiT-XL & 0.9 & 0.1 & 1e-2 $\rightarrow$ 0 & 1.0 & LoRA-1:2 \\
       DiT-D7 & AdamW(lr=2e-4, wd=0.0) & $\eta_{\text{min}}=1\text{e-4}$ & DiT-D14 & 0.9 & 0.1 & 1e-2 $\rightarrow$ 0 & 1.0 & LoRA-1:2 \\
       SiT-D14 & AdamW(lr=2e-4, wd=0.0) & $\eta_{\text{min}}=1\text{e-4}$ & SiT-XL & 0.9 & 0.1 & 2e-4 $\rightarrow$ 0 & 1.0 & LoRA-1:2 \\
       MAR-D16 & AdamW(lr=2e-4, wd=0.0) & $\eta_{\text{min}}=1\text{e-4}$ & MAR-Large & 0.9 & 0.1 & 1e-2 $\rightarrow$ 0 & 1.0 & LoRA-1:2 \\
    \bottomrule
    \end{tabular}
    %}
    \caption{Training details and hyper-parameters for mask training}
    \label{tab:appendix_hyperparam}
\end{table*}

\section{Visualization of Pruning Decisions}

Figures \ref{fig:decision-1-2}, \ref{fig:decision-2-4} and \ref{fig:decision-7-14} visualize the dynamics of pruning decisions during training for the 1:2, 2:4, and 7:14 pruning schemes. Different divisions lead to varying search spaces, which in turn result in various solutions. For both the 1:2 and 2:4 schemes, good decisions can be learned in only one epoch, while the 7:14 scheme encounters optimization difficulty. This is due to the $\binom{14}{7}$=3,432 candidates, which is too huge and thus cannot be adequately sampled within a single epoch. Therefore, in practical applications, we use the 1:2 or 2:4 schemes for learnable layer pruning.

\section{Details of Masked Knowledge Distillation}

\paragraph{Training Loss.} This work deploys a standard knowledge distillation to learn a good student model by mimicking the pre-trained teacher. The loss function is formalized as:
\begin{equation}
    \mathcal{L} = \alpha_{\text{KD}} \cdot \mathcal{L}_{\text{KD}} + \alpha_{\text{Diff}} \cdot \mathcal{L}_{\text{Diff}} + \beta \cdot \mathcal{L}_{\text{Rep}}
    \label{eqn:repkd}
\end{equation}
Here, $\mathcal{L}{\text{KD}}$ denotes the Mean Squared Error between the outputs of the student and teacher models. $\mathcal{L}{\text{Diff}}$ represents the original pre-training loss function. Finally, $\mathcal{L}_{\text{Rep}}$ corresponds to the masked distillation loss applied to the hidden states, as illustrated in Figure~\ref{fig:kd}, which encourages alignment between the intermediate representations of the pruned model and the original model. The corresponding hyperparameters $\alpha_{\text{KD}}$, $\alpha_{\text{Diff}}$ and $\alpha_{\text{Rep}}$ can be found in Table~\ref{tab:appendix_hyperparam}. 

\paragraph{Hidden State Alignment.} The masked distillation loss $\mathcal{L}_{\text{Rep}}$ is critical for aligning the intermediate representations of the student and teacher models. During the recovery phase, each layer of the student model is designed to replicate the output hidden states of a corresponding two-layer local block from the teacher model. Depth pruning does not alter the internal dimensions of the layers, enabling direct alignment without additional projection layers. For models such as SiTs, where hidden state losses are more pronounced due to their unique interpolant-based architecture, a smaller coefficient $\beta$ is applied to $\mathcal{L}_{\text{Rep}}$ to mitigate potential training instability. The gradual decrease in $\beta$ throughout training further reduces the risk of negative impacts on convergence.

\paragraph{Iterative Pruning and Distillation.} Table~\ref{tab:teacher_choice} assesses the effectiveness of iterative pruning and teacher selection strategies. To obtain a TinyDiT-D7, we can either directly prune a DiT-XL with 28 layers or craft a TinyDiT-D14 first and then iteratively produce the small models. To investigate the impact of teacher choice and the method for obtaining the initial weights of the student model, we derived the initial weights of TinyDiT-D7 by pruning both a pre-trained model and a crafted intermediate model. Subsequently, we used both the trained and crafted models as teachers for the pruned student models. Across four experimental settings, pruning and distilling using the crafted intermediate model yielded the best performance. Notably, models pruned from the crafted model outperformed those pruned from the pre-trained model regardless of the teacher model employed in the distillation process. We attribute this superior performance to two factors: first, the crafted model's structure is better adapted to knowledge distillation since it was trained using a distillation method; second, the reduced search space facilitates finding a more favorable initial state for the student model.

\begin{table}[t]
    \centering
    \resizebox{\linewidth}{!}{
    \begin{tabular}{l c| c c c c c l}
    \toprule
        \bf Teacher Model & \bf Pruned From & \bf IS & \bf FID & \bf sFID & \bf Prec. & \bf Recall \\
        \midrule
       % DiT-XL/2 & DiT-XL/2 & 1.31 & 370.56 & 143.30 & 0.0030 & 0.0000  \\
        DiT-XL/2 & DiT-XL/2 & 29.46 & 56.18 & 26.03 & 0.43 & 0.51  \\
        %DiT-XL/2 & TinyDiT-D14 & 2.60 & 275.53 & 158.71 & 0.0225 & 0.0011   \\
        DiT-XL/2 & TinyDiT-D14 & 51.96 & 36.69 & 28.28 & 0.53 & 0.59   \\
        % TinyDiT-D14 & DiT-XL/2 & 1.32 & 373.06 & 143.71 & 0.0027 & 0.0000  \\ %[1, 4, 9, 15, 18, 23, 24]
        TinyDiT-D14 & DiT-XL/2 & 28.30 & 58.73 & 29.53 & 0.41 & 0.50 \\ %[1, 4, 9, 15, 18, 23, 24]
        %TinyDiT-D14 & TinyDiT-D14 & 2.68 & 284.08 & 168.98 & 0.0159 & 0.0000  %\\ %[0, 2, 4, 7, 9, 11, 13]"
        TinyDiT-D14 & TinyDiT-D14 & 57.97 & 32.47 & 26.05 & 0.55 & 0.60  \\ %[0, 2, 4, 7, 9, 11, 13]"
        
    \bottomrule
    \end{tabular}   
    }
    \caption{TinyDiT-D7 is pruned and distilled with different teacher models for 10k, sample steps is 64, original weights are used for sampling rather than EMA.}
    \label{tab:teacher_choice}
\end{table}

\section{Analytical Experiments}

\begin{figure}
    \centering
    \includegraphics[width=\linewidth]{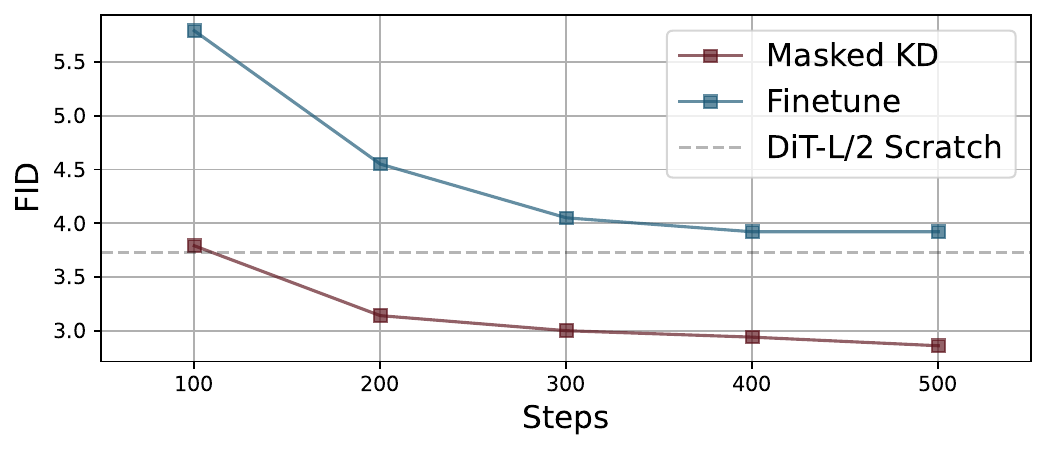}
    \caption{FID and training steps.}
    \label{fig:fid_steps}
\end{figure}

\paragraph{Training Strategies} 
Figure~\ref{fig:fid_steps} illustrates the effectiveness of standard fine-tuning and knowledge distillation (KD), where we prune DiT-XL to 14 layers and then apply various fine-tuning methods. Figure 3 presents the FID scores across 100K to 500K steps. It is evident that the standard fine-tuning method allows TinyDiT-D14 to achieve performance comparable to DiT-L while offering faster inference. Additionally, we confirm the significant effectiveness of distillation, which enables the model to surpass DiT-L at just 100K steps and achieve better FID scores than the 500K standard fine-tuned TinyDiT-D14. This is because the distillation of hidden layers provides stronger supervision. Further increasing the training steps to 500K leads to significantly better results.

\paragraph{Learning Rate.} We also search on some key hyperparameters such as learning rates in Table \ref{tab:lr}. We identify the effectiveness of lr=2e-4 and apply it to all models and experiments.

\begin{table}[t]
    \centering
    \resizebox{\linewidth}{!}{
    \begin{tabular}{l|c c c c c}
    \toprule
        Learning Rate & IS & FID & sFID & Prec. & Recall \\
        \midrule
        lr=2e-4 & 207.27 & 3.73 & 5.04 & 0.8127 & 0.5401 \\
        lr=1e-4 & 194.31 & 4.10 & 5.01 & 0.8053 & 0.5413  \\
        lr=5e-5 & 161.40 & 6.63 & 6.69 & 0.7419 & 0.5705 \\
    \bottomrule
    \end{tabular}
    }
    \caption{The effect of Learning rato for TinyDiT-D14 finetuning w/o knowledge distillation}
    \label{tab:lr}
\end{table}

\section{Visulization}

\begin{figure*}[t]
    \centering
    \includegraphics[width=0.60\linewidth]{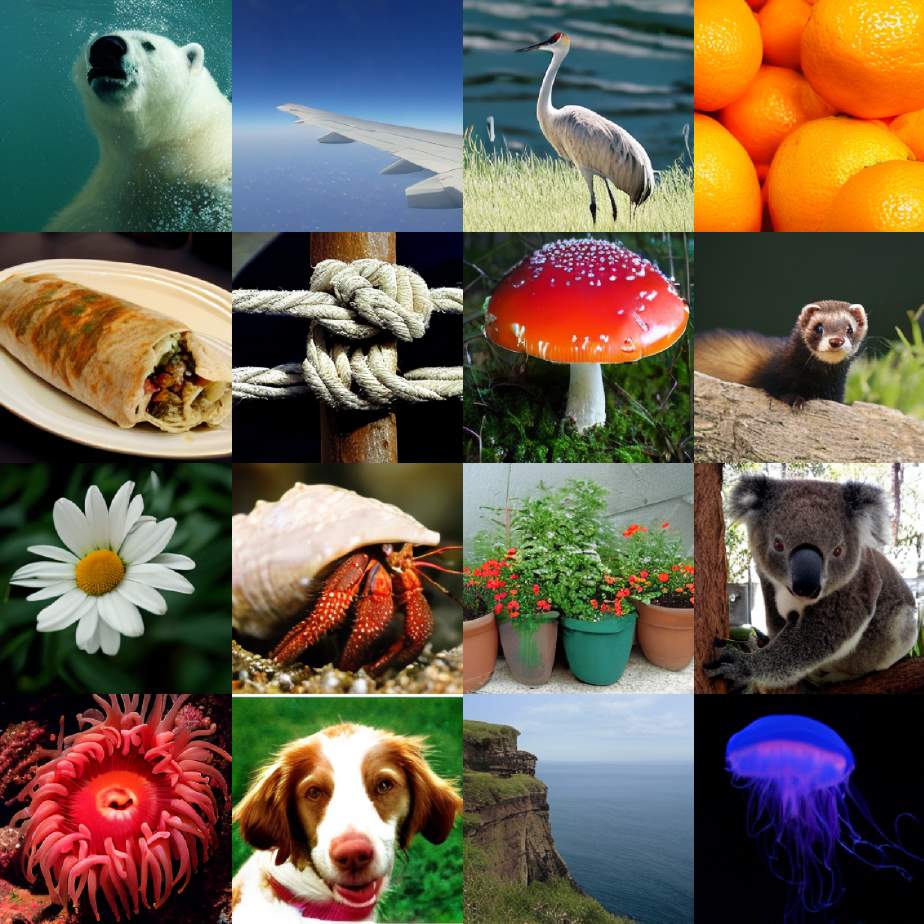}
    \caption{Generated images from TinySiT-D14}
    \label{fig:sit_vis}
    
    \includegraphics[width=0.60\linewidth]{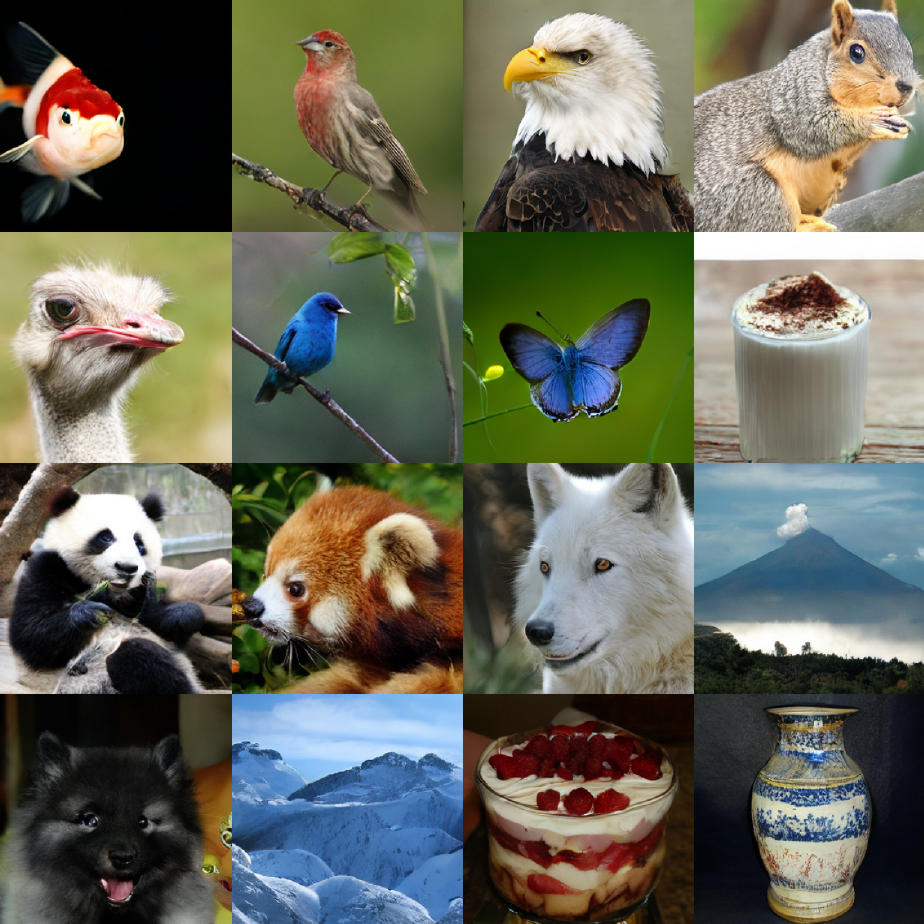}
    \caption{Generated images from TinyMAR-D16}
    \label{fig:mar_vis}
\end{figure*}

Figure \ref{fig:sit_vis} and \ref{fig:mar_vis}  showcase the generated images from TinySiT-D14 and TinyMAR-D16, which were compressed from the official checkpoints. These models were trained using only 7\% and 10\% of the original pre-training costs, respectively, and were distilled using the proposed masked knowledge distillation method. Despite compression, the models are capable of generating plausible results with only 50\% of depth.

\section{Limitations}

In this work, we explore a learnable depth pruning method to accelerate diffusion transformer models for conditional image generation. As Diffusion Transformers have shown significant advancements in text-to-image generation, it is valuable to conduct a systematic analysis of the impact of layer removal within the text-to-image tasks. Additionally, there exist other interesting depth pruning strategies that need to be studied, such as more fine-grained pruning strategies that remove attention layers and MLP layers independently instead of removing entire transformer blocks. We leave these investigations for future work.

\clearpage

\end{document}